\documentclass[10pt,twocolumn,letterpaper]{article}
    
\usepackage{cvpr}
\usepackage{times}
\usepackage{epsfig}
\usepackage{graphicx}
\usepackage{amsmath}
\usepackage{amssymb}
\usepackage[]{algorithm2e} %For algorithm block

\usepackage{soul}
\usepackage{authblk}
\usepackage[pagebackref=true,breaklinks=true,letterpaper=true,colorlinks,bookmarks=false]{hyperref}

\cvprfinalcopy % *** Uncomment this line for the final submission

\def\cvprPaperID{1253} % *** Enter the CVPR Paper ID here

\ifcvprfinal\fi
\begin{document}

%%%%%%%%% TITLE
\title{3D-A-Nets: 3D Deep Dense Descriptor for Volumetric Shapes with Adversarial Networks }
\author{Mengwei Ren}
\author{Liang Niu}
\author{Yi Fang\thanks{Address: 5 Metrotech Center LC024, Brooklyn, NY, 11201; Email: yfang@nyu.edu; Tel: +1-646-854-8866}}
\affil{NYU Multimedia and Visual Computing Lab}
\affil{NYU Abu Dhabi, UAE}
\affil{New York University, Brooklyn, NY, USA}
% \author{\textbf{Mengwei Ren, Liang Niu, Yi Fang\thanks{yfang@nyu.edu}}\\
% NYU Multimedia and Visual Computing Lab\\
% Department of Electrical and Computer Engineering, NYU Abu Dhabi, UAE\\
% Department of Electrical and Computer Engineering, NYU Tandon School of Engineering, USA\\
% Department of Computer Science and Engineering, NYU Tandon School of Engineering, USA\\
\maketitle
%\thispagestyle{empty}

%%%%%%%%% ABSTRACT
\begin{abstract}Recently researchers have been shifting their focus towards learned 3D shape descriptors from hand-craft ones to better address challenging issues of the deformation and structural variation inherently present in 3D objects. 3D geometric data are often transformed to 3D Voxel grids with regular format in order to be better fed to a deep neural net architecture. However, the computational intractability of direct application of 3D convolutional nets to 3D volumetric data severely limits the efficiency (i.e. slow processing) and effectiveness (i.e. unsatisfied accuracy) in processing 3D geometric data. In this paper, powered with a novel design of adversarial networks (3D-A-Nets), we have developed a novel 3D deep dense shape descriptor (3D-DDSD) to address the challenging issues of efficient and effective 3D volumetric data processing.
% Particularly, in this research,
We developed new definition of 2D multilayer dense representation (MDR) of 3D volumetric data to extract concise but geometrically informative shape description and a novel design of adversarial networks that jointly train a set of convolution neural network (CNN), recurrent neural network (RNN) and an adversarial discriminator.
% for the robust 3D-DDSD for volumetric shapes.
More specifically, the generator network produces 3D shape features that encourages the clustering of samples from the same category with correct class label, whereas the discriminator network discourages the clustering by assigning them misleading adversarial class labels. By addressing the challenges posed by the computational inefficiency of direct application of CNN to 3D volumetric data, 3D-A-Nets can learn high-quality 3D-DSDD which demonstrates superior performance on 3D shape classification and retrieval over other state-of-the-art techniques by a great margin.
\end{abstract}

%%%%%%%%% BODY TEXT

\section{Introduction}
\subsection{Background}
With recent advancements in range sensors (i.e. LiDAR and RGBD cameras) and imaging technologies (i.e. 3D MRI), the amount of available 3D geometric data has explosively increased in a variety of applications such as engineering, entertainment, and medicine \cite{Albarelli2011A, rodola2012game, bronstein2006efficient, chen2009benchmark, de2008hierarchical, Katz2005Mesh, osada2002shape, rustamov2007laplace}. It is therefore of great interests to develop methods that can automatically analyze the large amount of 3D geometric data for different tasks (e.g., 3D object recognition, classification and retrieval). To that end, a lot of effort have been made, many of which have been focused on building robust 3D shape representations. However, the structural variation of 3D objects (i.e. 3D human models with different poses, 3D car models with various design pattern) pose great challenges on learning a high-quality 3D shape descriptor. Recently, the deep learning techniques (i.e. convolutional networks (CNNs)) have demonstrated significant improvement of the performance in an 2D image object recognition \cite{lin2017focal, he2016deep, redmon2016yolo9000,shi2017detecting,bai2017scalable} as well as the efficiency and effectiveness in other computer vision tasks \cite{he2017mask, wang2017fast,Hegde2016FusionNet}. The promising performance of deep neural networks motivates 3D computer vision researchers transform the 3D geometric data to 3D Voxel grids (volumetric shapes) in regular format so that the 3D data can be fed to a deep net architecture \cite{Wu20153D} for further processing. While the extension of the deep learning to volumetric shapes is conceptually simple, the computational intractability of direct application of 3D convolutional nets to 3D data severely limits the efficiency (i.e. slow processing) and effectiveness (i.e. unsatisfied accuracy) in processing 3D geometric data for object recognition \cite{Maturana2015VoxNet}. It is reported in \cite{hinton2012does} that the computational intractability often caused ``nightmare'' for 3D convolutional nets in representing 3D volumetric shapes. We will review recent related works in deep neural networks including CNNs and RNNs, deep learning for 3D shape descriptor for object recognition as well as adversarial networks.   

%% 1.2. Related works
\subsection{Related Works}
Learning a deformation-invariant shape descriptor from a large collection of 3D geometric data with significant structural variation is of great importance as the quality of the shape descriptor ultimately determines the 3D object recognition. Related works mainly include three parts: 1) hand-crafted 3D shape descriptors, 2) data-driven shape feature learning, 3) adversarial networks.

% 1.2.1 Hand-Crafted 3D shape Descriptors
\subsubsection{Hand-Crafted 3D shape Descriptors}

3D shape descriptor are succinct and compact representations of 3D object that capture the geometric essence of a 3D object. Some existing shape descriptors have been developed to describe the 3D objects \cite{Hegde2016FusionNet,Wu20153D,Maturana2015VoxNet}. The earlier D2 shape distribution, statistical moments, Fourier descriptor, Light Field Descriptor, Eigenvalue Descriptor have been proposed to describe the 3D shape particularly for rigid 3D objects. The Spin Image \cite{zeng20163dmatch} was developed based on the dense collection of 3D points and surface normals. There are also few histograms (i.e. geometry histogram in \cite{zeng20163dmatch}), feature histogram \cite{zeng20163dmatch} and signatures of histograms) shape descriptors developed based on the distribution of a type of statistical geometric properties. The efforts on robust 3D shape feature are further developed by heat diffusion geometry. A global shape descriptor, named temperature distribution (TD) descriptor, is developed based on HKS information at a single scale to represent the entire shape \cite{fang2011temperature}. Hand-crafted shape descriptors are often not robust enough to deal with structural variations and incompleteness present in 3D real-world models, and are often not able to be generalized to data of different modality. Discriminative feature learning from large datasets provides an alternative way to construct deformation-invariant features. 

\subsubsection{Shape feature learning}

The bag-of-features (BOF) is first introduced to learn to extract a frequency histogram of geometric words for shape retrieval \cite{eitz2012humans,eitz2011sketch,hu2013performance}. To learn global features, \cite{Fang20153D} adopted auto-encoder with the distribution of HKS learns a deformation-invariant shape descriptor. Recent advancement in deep learning motivates researchers to learn a 3D shape descriptor from a large-scale dataset using deep neural networks. However, to feed the 3D geometric data to neural networks, the 3D geometric data are often transformed to 3D Voxel grids or a collections of 2D projection images from different views. 

\textbf{Voxelization-based methods:} The volumetric representation plays an important role in computer graphics community since the 1980s. It provides a uniform, simple and robust description to synthetic and measured objects and founds the basis of volume graphics \cite{Kaufman1993Volume}. In other words, voxel is an extension of pixel, and the binary volume is an extension of binary image. Recently, many researchers begin to develop 3D CNN on volumetric shapes. \cite{Wu20153D} voxelizes the 3D shape into 3D grids and train a generative model for 3D shape recognition using convolutional deep belief network. Similarly, \cite{Maturana2015VoxNet} proposed a real-time 3D supervised learning architecture on volumetric 3D shapes. Apart from supervised CNN, \cite{Wu2016Learning} generate 3D objects from a probabilistic space by leveraging advances in volumetric convolutional networks and generative adversarial nets, and the unsupervisedly learned features can be widely used in 3D object recognition. Besides, \cite{Sharma2016VConv} proposed a 3D convolutional auto-encoder for recognizing 3D shape. 

\textbf{Projection-based methods:} Differ from the direct 3D representation, the 3D shapes can also be projected to 2D space. \cite{Su2015Multi} proposed a multi-view CNN for 3D shape recognition by using CNN to extract visual features from images with different views, and employing max-pooling across views to learn one compact shape descriptor. The LFD \cite{Chen2003On} extract features from the light fields rendered from cameras on a sphere exhaustively to improve the robustness against rotations. By the same means, \cite{Bai20153D} proposed a coding framework for constructing a compact descriptor based on a set of 2D views in the format of depth buffer rendered from each 3D object.

\begin{figure*}[h]
\begin{center}
\includegraphics[width=0.9\linewidth]{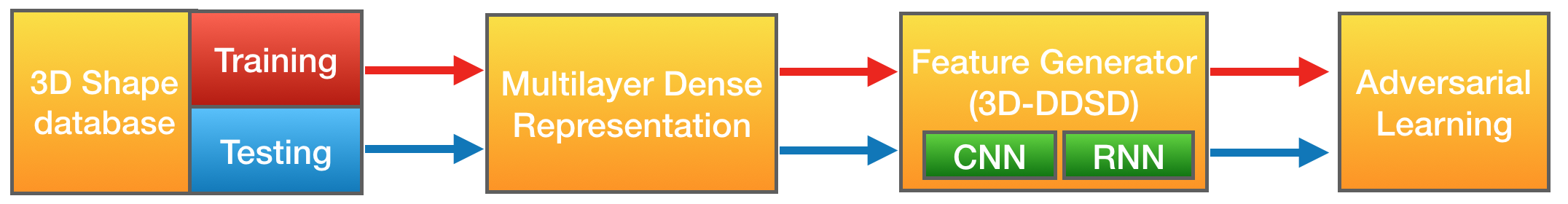}
\end{center}
\caption{Four main components of the proposed method: 1) 3D shape database; 2) Generation of MDR; 3) Feature generator; 4) Adversarial learning on feature level.}
\label{fig:Main Components}
\end{figure*}

\begin{figure*}[h]
\begin{center}
    \includegraphics[width=0.9\linewidth]{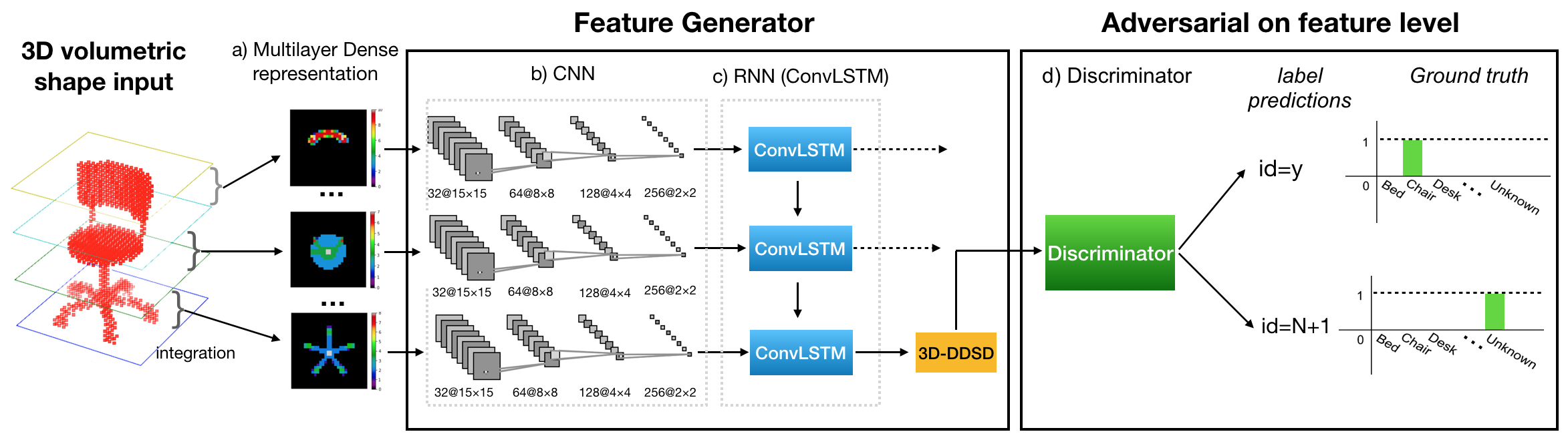}
\end{center}
\caption{Overview of the proposed approach. The feature generator takes MDR (transformed from 3D shape) as input, and generate 3D-DDSD; The discriminator takes 3D-DDSD as input, and produces class label (1 to N represents N predefined class labels, and N+1 represents `adversarial') }
\label{fig:pipeline}
\end{figure*}

\subsubsection{Adversarial networks}
A classifier (i.e. deep neural network) unfortunately starts losing the incentive for better recognizing an object with a correct class label when it becomes more widely trained and deployed \cite{lowd2005adversarial}. Recent studies \cite{goodfellow2014generative,lowd2005adversarial} have shown that adversarial training, where a set of neural networks jointly learn together by pursuing competing goals, can help classifier dramatically improve the recognition performance benefiting from learning with more adversarial examples that were intentionally designed to cause the classifier to make mistakes. One promising technique that trains the deep neural network with adversarial loss is Generative Adversarial Networks (GANs), which has recently received great popularity due to its ability to learn deep feature representations without extensively annotated/labeled training data. GANs was proposed to train generative models in an adversarial process where two networks (generator network and discriminator network) are modeled as two players in a min-max game. A generator network is learning to produce synthetic data to resemble real ones and a discriminator network distinguish between the generator’s output and true data. The GANs have been successfully applied in various applications such as 3D model synthesis \cite{3dgan}, 2D image synthesis \cite{wang2017fast}, image super resolution \cite{ledig2016photo}, style transfer \cite{zhu2017unpaired,taigman2016unsupervised} and etc.
In addition, recent works of adversarial learning at feature level rather than object level has shown its  promising capability of discriminative feature learning.
% Recent Fader Networks \cite{lample2017fader} demonstrates the compete at feature level opens up a new direction towards robust feature learning with adversarial setting. 
% Recent work in image attributes isolation has indicated the robustness of adversarial learning in the feature space rather than image space, for example, Fader Networks \cite{lample2017fader} shows a discriminator which isolates attributes from features generated from generator.
Recent development of Fader Networks \cite{lample2017fader}, with Generator and Discriminator competing in the latent space opens up new direction towards robust adversarial feature learning.
Given the ability to learn discriminative representation, we propose a novel framework, named 3D Deep Dense Descriptor Learning with Adversarial Networks (3D-A-Net), to learn a deformation-invariant 3D shape descriptor with an adversarial process as described below.

% 1.3 Our solution
\subsection{Our solution: 3D-A-Nets towards 3D-DDSD}

Inspired by this new direction of adversarial feature learning, 
this paper proposed to learn 3D shape adversarial features for robust 3D recognition and retrieval.
This paper will focus on 1) development of new multilayer dense data representation (MDR) of 3D object to reduce the computational load for more efficient processing, and 2) development of more effective network architecture (3D-A-Nets) via a novel design of deep adversarial networks that jointly train a set of convolution neural network (CNN), recurrent neural network (RNN) and an adversarial discriminator for the robust 3D deep dense shape descriptor (3D-DDSD) for volumetric shapes.
Figure \ref{fig:Main Components} and Figure \ref{fig:pipeline} illustrate the pipeline of the proposed method. There are four main components. The first component is a 3D shape database where a large number of 3D volumetric data are stored. The second component is generation of multilayer dense representation of 3D volumetric where each slice of representation is an integration of multi-segments along one particular axis.
The third component is the feature generator which consists of a CNN and a many-to-one ConvLSTM \cite{Shi2015Convolutional}. The CNN is used to extract the geometric feature in each slice whereas the ConvLSTM is used to explore the spatial information across different slices. The 3D shape feature produced by generator in this component are learnt towards the clustering of samples from same category. Whereas the fourth component, a discriminator, is trained to assign the generated feature with incorrect adversarial class labels.
% the generator network produces 3D shape features that encourages the clustering of samples from the same category with correct class label, whereas the discriminator network discourages the clustering by assigning them incorrect adversarial class labels
There are two communication routes in the pipeline, in which the route along the red arrow is utilized as jointly train a set of CNN, RNN and an adversarial discriminator for the learning of 3D deep dense shape descriptor. The route along the blue arrow is for the testing data. After training, feature generator is used to produce the 3D deep dense shape descriptor (3D-DDSD).

%% 2. Method
\section{Method}
\subsection{Generative Adversarial Network}
In Generative Adversarial Network (GAN), a two-player minimax game is played between two opponents: Generator $G$ and Discriminator $D$. $G$ tries to generate a realistic image $G(z)$ from a random noise $z$ in order to fool $D$, whereas $D$ tries to distinguish the synthesized image $G(z)$ from real image $x$. To this end, $G$ and $D$ are trained alternatively to optimize the following objectives respectively:

\begin{equation}
\begin{aligned}
\underset{G}{max}V_G(D,G)=& E_{z\sim p_z(z)}[\log(D(G(z)))]\\
\underset{D}{max}V_D(D,G)=& E_{x\sim p_d(x)}[\log D(x)]
\\ & +E_{z\sim p_z(z)}[\log(1-D(G(z)))]\\
\end{aligned}
\end{equation}

It is proved in \cite{goodfellow2014generative} that when the game achieves the global optimum, the distinguishing precision of $D$ stays at around 50\%, and $G$ is able to synthesize authentic-looking images, which have identical data distributions to the real training samples. By applying GANs, the generation performance will be significantly strengthened benefiting from the adversarial competing training process.  

\subsection{Proposed Model: 3D-A-Nets}
The proposed network, 3D Adversarial Network (3D-A-Nets), consists of a feature generator $G$, which tries to learn the distinguishable 3D-DDSD from the input MDR, and a feature-level discriminator $D$, pursuring the competing goals with $G$, which attempts to classify all its input 3D-DDSD towards the `adversarial' class, indicating the 3D-DDSD is too ambiguous to be assigned with an explicit class label.
Compared with vanilla GANs and other 3D shape related GANs, three major novelties of 3D-A-Nets are as follows. Firstly, the input of $G$ is the newly defined Multilayer Dense Representation (MDR) instead of a random noise $z$. Compared with 3D voxel volume, MDR is concise but geometrically informative to be better fed into a Neural Network (NN). In addition, to deal with the spatial relationship across slice s in MDR, Recurrent Neural Network (RNN), specifically, Convolutional Long Short Term Memory (ConvLSTM), is introduced into $G$. RNN takes the feature map sequence extracted by CNN as input, and outputs the 3D-DDSD as a latent space variable (shown in Figure \ref{fig:pipeline}). Third, adversarial learning is applied, which encourages $G$ to learn to generate a more distinguishable 3D-DDSD.

\subsubsection{Problem Formulation}
Given a 3D volumetric shape $x$ with label $y^d$, where $d \in \{1, 2, \ldots, N\}$, and $N$ refers to the total number of shape classes, the end goal is to learn a model which is able to generate a robust 3D-DDSD for any input shape $x$, as well as to estimate the correct label $y^d$ of $x$ with satisfying accuracy. Our approach is to jointly train a feature generator $G$ and an adversarial discriminator $D$ with the architecture illustrated in Figure \ref{fig:pipeline}. Specifically, the input is first transformed to the multilayer dense representation (see (a) in Figure \ref{fig:Main Components}), then the 3D-DDSD is learned from the CNN-RNN fused feature generator $G$ (see (b) in Figure \ref{fig:Main Components}). The discriminator $D$ takes the 3D-DDSD, and attempts to classify its input as an `adversarial' class $y^{N+1}$, indicating that the shape descriptor is too inexplicit to be assigned with an adversarial label. In the adversarial process, $G$ learns to encode the multilayer representation to a robust 3D-DDSD in order to push the $D$ to make a right estimation. 

\subsubsection{Multilayer Dense Representation (MDR)}
\begin{figure}[h]
\begin{center}
    \includegraphics[width=0.9\linewidth]{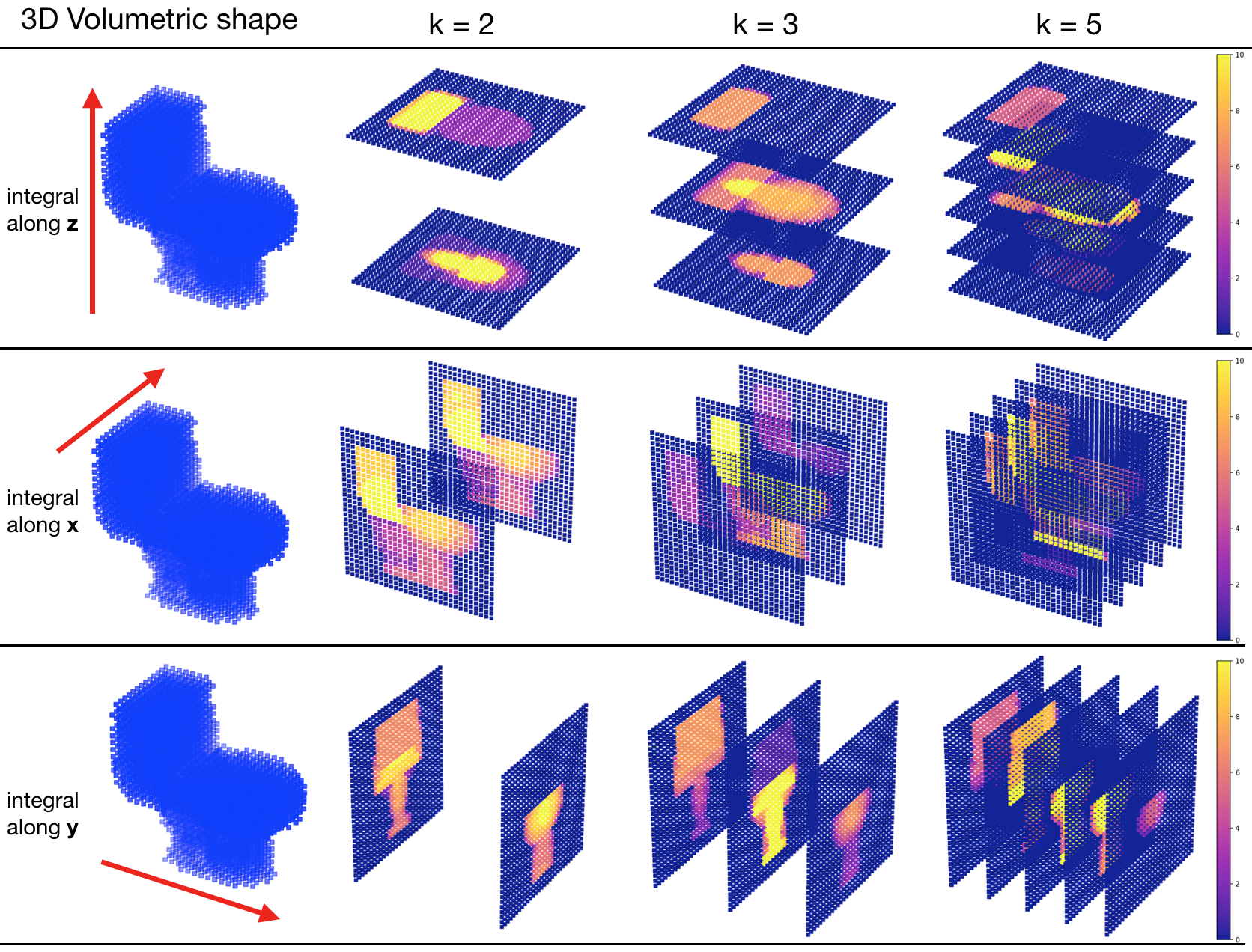}
\end{center}
\caption{MDRs of a `toilet' shape with different parameter settings. The left column displays the 3D volumetric shapes, and the following three columns illustrate the MDR $L^k$ with $k=2,3,5$, respectively. Each row shows the multi-segments integration along different axis ($z,x,y$).}
\label{fig:MDR}
\end{figure}

Multilayer Dense Representation (MDR) is defined as an organized 2D dense representation sequence, in which each element is computed by stacking continuous segments in the 3D volumetric data $x$ along specified axis. Compared with direct application of 3D CNN to 3D volumetric shape, employing 2D CNN to MDR addresses the computational intractability while maintaining geometrical information in 3D object recognition task. For a given 3D shape $x$, MDR algorithm $S: x \rightarrow L^k$ is applied to pack the 3D shape $x$ to a series of 2D dense representations, where $L$ is a series of 2D representations with $k$ elements. The MDR obtained from a 3D shape is then organized as a spatiotemporal sequence, and is fed into CNN-RNN fused network for feature generation. Detailed algorithm is explained in Algorithm \ref{algo:MDR}.
\begin{algorithm}[h]
    \KwData{3D Shape x with resolution $30 \times 30 \times 30$}
    \KwResult{Multilayer Dense Representation $L^k$}
    initialization \;
    $i \leftarrow 0$;\\
    $n \leftarrow $ 30/k ;\\
    \While{$i \neq k$}{
    $L^i \leftarrow $ segment integral of $x$ along axis $ax$ from $ax=i*n$ to $ax=(i+1)*n$\\
    $i \leftarrow i+1$
    }
    \caption{Multilayer Dense Representation Generation Algorithm}
    \label{algo:MDR}
\end{algorithm}

As illustrated in Figure \ref{fig:MDR}, three MDRs of a sample volumetric shape `toilet' are compared. The first column displays the 3D shape with resolution $30 \times 30 \times 30$, and the following three columns demonstrate the MDR $L^2$, $L^3$, $L^5$, respectively. Each MDR is computed along three axis $x,y,z$, as shown per row of the figure. 
\begin{figure*}[h]
\begin{center}
\includegraphics[width=0.9\linewidth]{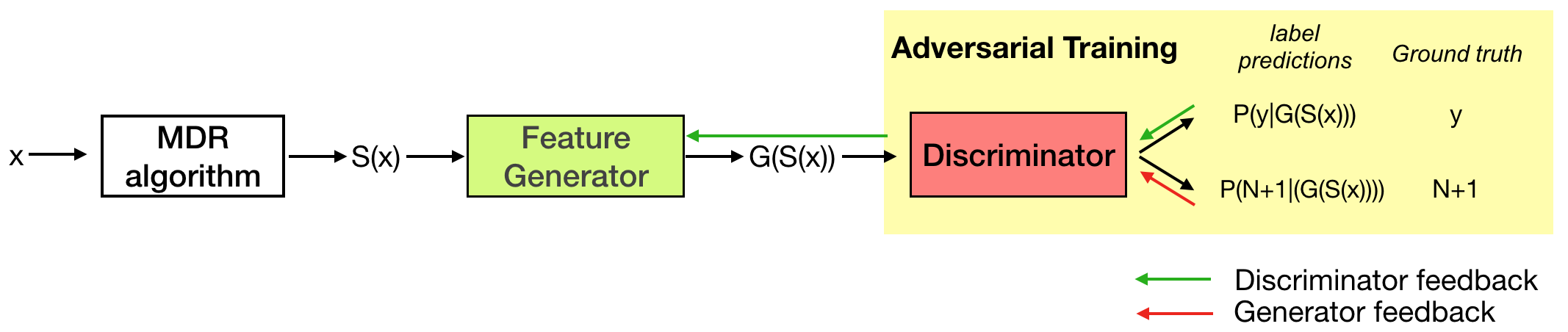}
\end{center}
\caption{Data Flow Pipeline of 3D-A-Nets. $x$ is the input voxel volume 3D data, which is then transformed to Multilayer Dense Representation $S(x)$. Then CNN-RNN fused Feature Generator takes $S(x)$ as input and outputs 3D-DDSD $G(S(x))$. Generated 3D-DDSD is sent to Discriminator for adversarial training (as shown in the yellow region). The forward data flow is drawn in black arrows, and backward feedback is drawn in green and red.}
\label{fig:3D-A-Net}
\end{figure*}

\subsubsection{Learning 3D-DDSD from CNN-RNN fused feature generator}
To learn a deep feature from the input MDR, a CNN-RNN (ConvLSTM) fused feature generator \noindent $G_{\theta_{gen}}: S(x) \rightarrow R$ is applied, where $S(x)$ is the MDR of 3D shape $x$, $\theta_{gen}$ are parameters of CNN and RNN, and $G_{\theta_{gen}}(S(x))$ is the desired 3D-DDSD. Specifically, CNN is used to extract feature map for each slice in $S(x)$, consisting a feature sequence of the MDR. Considering the spatiotemporal relationship of the sequence, RNN is well adapted in transforming the sequence to a single latent variable, which describes the 3D volumetric shape. In the proposed 3D-A-Nets, Convolutional LSTM (ConvLSTM) is applied, which is formally defined as:
\begin{equation}
\begin{array}{ll}
    i_t = \sigma(W_{xi}\ast L_t + W_{hi}\ast H_{t-1} + W_{ci}\circ C_{t-1} + b_i)\\
    f_t = \sigma(W_{xf}\ast L_t + W_{hf}\ast H_{t-1} + W_{cf}\circ C_{t-1} + b_f)\\
    C_t = f_t \circ C_{t-1} + i_t \circ \tanh(W_{xc} \ast L_t + W_{hc} * H_{t-1} + b_c)\\
    o_t = \sigma(W_{xo} * L_t + W_{ho}\ast H_{t-1} + W_{co} \circ C_t + b_o\\
    H_t = o_t \circ \tanh(C_t)
\end{array}
\end{equation}
where '$\ast$' is the convolution operator and '$\circ$' denotes the Handamard product. $i_t, f_t, o_t$ represent input gate, forget gate, output gate respectively; $L_1, L_2, \ldots ,L_t$ are the sequential inputs. Specifically, each $L$ is a 2D slice in MDR, represented by a $N \times N$ grid. Each cell of the grid is a stacked dense value $D$. Thus tensor $L$ can be further defined as $ L\in\mathbf{R}^{ N \times N \times D}$. If $k$ segments are acquired, a sequence of tensors $L_1, L_2, \ldots ,L_k$ is generated. Cell outputs $C_1,C_2,\ldots,C_t$ and hidden states $H_1,H_2,\ldots,H_t$ are also 3D tensors. Here, $C_t$ is the desired 3D Deep Dense shape descriptor. 

\subsubsection{Adversarial Learning Objectives}
In object recognition, classification and retrieval, the inter-class variance of the deep learned shape descriptors is considered as a watershed of a robust 3D shape descriptor. To address the issue that a classifier starts losing the incentive for better recognizing an object with a correct label, we proposed to introduce a novel adversarial learning to encourage a more distinguishable 3D-DDSD from $G$, as well as a more sensitive classifier $D$, with the network structure demonstrated in Figure \ref{fig:3D-A-Net}.  

%Two important attributes of a good 3D shape descriptor are addressed in our consideration, purity and similarity.
%An important attribute of a robust 3D shape descriptor is addressed in our consideration, intra-class similarity.
% Purity means the descriptor should only contain information that is necessary to distinguish the shape.
%Similarity means the descriptor of shapes from same class should be very similar, it is the key of successfully performing classification or retrieval tasks on such descriptor. Thus we raised adversarial learning to help the model improve this attribute of generated DDSD.

$D$ takes a 3D-DDSD as input and estimates probabilities of a label vector $P_{\theta_{dis}}(y^{N+1}|G_{\theta_{gen}}(S(x)))$, where $\theta_{dis}$ are the discriminator's parameters.
% \textcolor{red}{
The competing process, the objective of the feature generator $G$ is to encode the $S(x)$ to a discriminative 3D-DDSD towards different 3D objects so that $D$ is able to predict its label without hesitation, whereas $D$ attempts to confuse the 3D-DDSD by trying to classify all its inputs to an additional class `adversarial' with label $N+1$, where $N$ is the number of total classes in the dataset, indicating that the input 3D-DDSD is too vague and inexplicit.
To this end, the objectives and loss functions are designed as follows to meet the requirements. 

%\textcolor{red}{Questions about loss function. Usually, for GAN, there should be both real samples and fakes samples to train discriminator. However, from the below loss function, only fake samples are used to train both generator and discriminator (feels a bit confused). Maybe Is it possible to use some predefined latent vector as ``real samples'' with G(S(x)) together to train discriminator?}

\begin{equation}
\begin{aligned}
\underset{G}{max}V_G(D,G) &= \\ E_{x,y\sim P_d(x,y)}&[\log P_{\theta_{dis}}(y^d|G_{\theta_{gen}}(S(x)))] \\
\underset{D}{max}V_D(D,G) &= \\ E_{x,y\sim P_d(x,y)}&[\log P_{\theta_{dis}}(y^{N+1}|G_{\theta_{gen}}(S(x)))]
\end{aligned}
\end{equation}

%$$\underset{G}{max}V_G(D,G) = E_{x,y\sim P_d(x,y)}[\log P_{\theta_{dis}}(y^d|G_{\theta_{gen}}(S(x)))]
%$$
%$$
%\underset{D}{max}V_D(D,G) = E_{x,y\sim P_d(x,y)}[\log P_{\theta_{dis}}(y^{N+1}|G_{\theta_{gen}}(S(x)))] 
%$$

%The objective of discriminator is to separate the input DDSD,???. And to compete with discriminator, $G$ is designed so it tries to make all 3D shape sunder the same class have very similar generated DDSD, $G_{\theta_{gen}}(S(x))$. Thus the objectives and loss functions are designed as follows to meet the design.

%The discriminator estimates probabilities of a label vector $P_{\theta_{dis}}(y^{N+1}|G_{\theta_{gen}}(S(x)))$, where $\theta_{dis}$ are the discriminator's parameters. The objective of $D$: 

%The objective of the feature generator $G$ is to encode the $S(x)$ to a discriminative DDSD towards different 3D objects.
%$P_{\theta_{dis}}(y^{d}|G_{\theta_{gen}}(S(x)))$
%$$\underset{G}{max}V_G(D,G) = E_{x,y\sim P_d(x,y)}[\log P_{\theta_{dis}}(y^d|G_{\theta_{gen}}(S(x)))]
%$$
\subsection{Comparison to prior voxelization-based methods}
We compare 3D-A-Nets with three recently developed 3D Neural Network based model variants for dealing with 3D shapes. 

\paragraph{3D ShapeNets}
3D ShapeNets uses a single Deep Belief Network (DBN) as its core architecture. Raw 3D data are input into the 5-layer DBN, including 3 convolutional layers, 1 fully connected layer and final layer with 4000 hidden units takes as input a combination of multinomial label variables and Bernoulli feature variables \cite{Wu20153D}. The authors propose to use binary variables on 3D voxel grid data, in which 1 presents empty space and 1 presents filled space or unknown space, so that the model could also taking 2.5 D data as input. This novelty improves the data type that can be taken by the model. However, 3D-A-Nets deploys MDR on input data and reaches better result in both effectiveness and efficiency comparing to 3D ShapeNets.
\paragraph{3D GAN}
3D GAN \cite{3dgan} is developed by Wu etc., as a feasible framework to extend vanilla 2D GAN to be able to generate 3D CAD format shape from noise $z$. In this work, the authors also raise a variant of 3D GAN that uses 3D VAE (Variational Autoencoder) as generator. This model provides the inspiration to deploy 3D convolutional and 3D deconvolutional layers when dealing with 3D data. Particularly, the 3D VAE architecture of generator is widely considered in many recent 3D shape related researches. However, it is a too large model that needs huge amount of GPU computation time and GPU memory when training, leads to a low efficiency even for low resolution 3D shapes. Recently researchers have been looking for a better model for 3D shape. Our 3D-A-Nets uses MDR to replace raw data and uses 2D CNN-RNN as major component, achieves better performance than models taking 3D raw data (CAD or point cloud or other formats) as input. 
\paragraph{VoxNet} VoxNet \cite{Maturana2015VoxNet} is a model developed by Daniel Maturana and Sebastian Scherer that takes  Volumetric Occupancy Grid, which is transformed from point cloud data, as input and the major component is a 5-layer 3D CNN network including 2 3D-Convolutional layers, one pooling layer and 2 fully connected layers. It finally outputs a $N$ length vector as the probabilistic estimate of label where $N$ is the number of classes. The authors addressed efficiency and effectiveness by introducing occupancy grid as input, however, the neural network architecture still uses 3D convolutional layers as major component so the computation time needed is still large. As shown in experiment result (See Table \ref{tab:comparison}), 3D-A-Nets outperforms VoxNet in classification accuracy. %and according to our experiments, the computation time of 3D-A-Nets needed for training and inference on same dataset also outperforms VoxNet???

\section{Experiments}
\begin{figure}[h]
\begin{center}
    \includegraphics[width=0.8\linewidth]{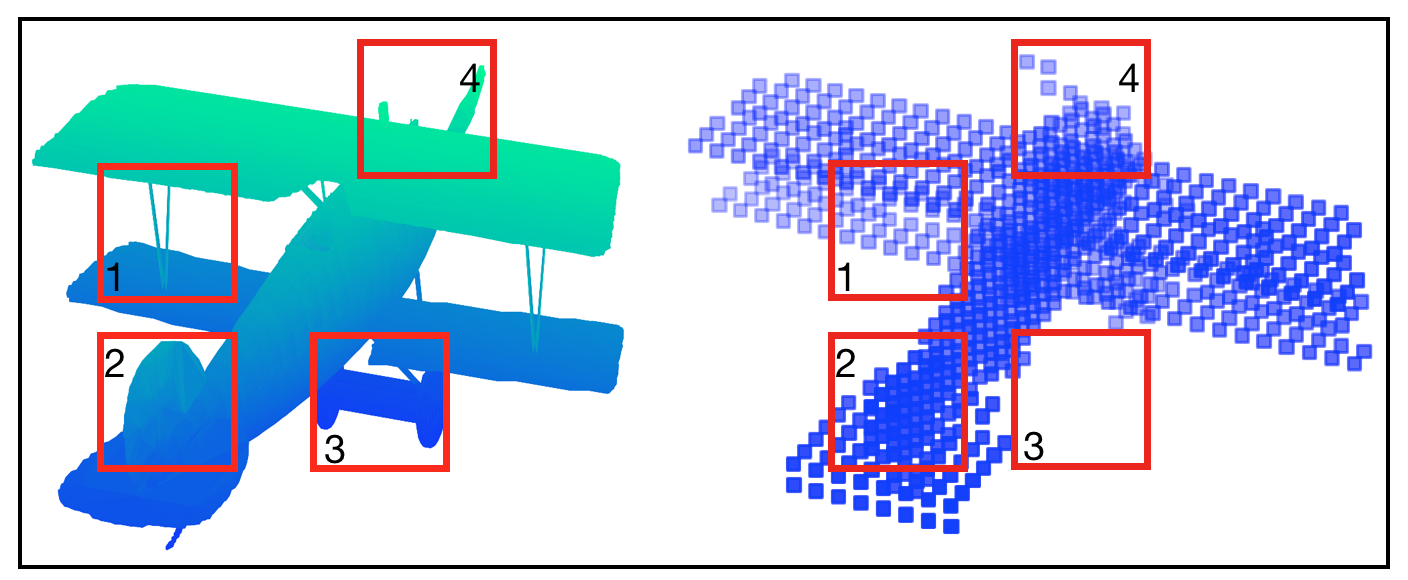}
\end{center}
\caption{Mesh (Left) and volumetric data (Right) comparison, with the corresponding detailed regions marked with red bounding box.)}
\label{fig:mesh_voxel}
\end{figure}

In this section, a set of experiments were carried out to evaluate performance of the proposed 3D-A-Nets via assessing the quality of the learned 3D-DDSD for shape classification and retrieval. The 3D models used in the experiments were chosen from the Princeton ModelNet40 \cite{Wu20153D} that contains 12,311 models and 40 categories with imbalance distribution of the number of models in each category. Our method is to compare against other three deep learning methods which were developed to process the 3D volumetric data. Please note that we currently did not compare 3D-A-Nets with projection-based methods that were developed based on using 2D projection views of a 3D model as input of the neural nets (i.e. MVCNN-MultiRes \cite{qi2016volumetric}, PANORAMA-NN \cite{3dor.20171045} etc.) for the reason: the 3D models (triangulated mesh format) that are used to produce 2D projection views to feed the above mentioned projection-based methods have much higher resolution than that of 3D models (3D voxel grids format) that were fed to our 3D-A-Nets and other voxelization-based method (i.e. 3D ShapeNets and VoxNet). To illustrate the resolution discrepancy, we display two 3D models with different type of data formats in Figure \ref{fig:mesh_voxel}. As shown in figure, the 3D airplane model on the left contains much more geometric details than its 3D voxel data on the right. The highlighted corresponding regions in red-box on both models clearly reflect the resolution discrepancy between two models. Therefore, given the severe loss of geometric details of 3D voxel data, it is not appropriate to have a direct comparison between voxelization-based and projection-based methods based on different data entry. We therefore compare 3D-A-Nets to the state-of-the-art methods using the same data format and leave the comparison to projection-based to future research (i.e. 2D images produced by projecting the 3D voxel model instead of high-resolution triangulated models).

\subsection{Implementation details}
\begin{figure}[h]
\begin{center}
    \includegraphics[width=1\linewidth]{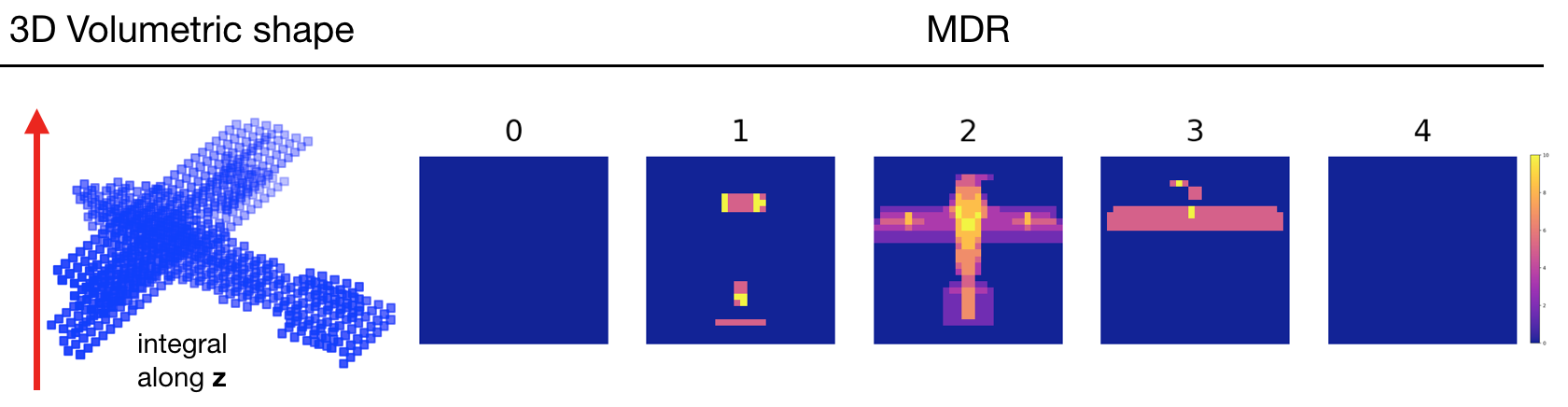}
\end{center}
\caption{MDR along axis $y$ of a sampled `airplane', with $k$=5. The first and last slice of MDR is empty (with all values equal zero)}
\label{fig:MDR_n5}
\end{figure}

%\textbf{Voxelization resolution}: In all experiments we use a fixed resolution $30 \times 30\times 30$ to maintain a balance between precision and speed. A $30 \times 30 \times 30$ volumetric grid has a decent estimation of the shape and the contour of the original mesh, while eliminating extra details. 
\textbf{Parameters setting of MDR}: In MDR algorithm, choosing an appropriate $k$, which is the number of slices in MDR, is crucial considering the trade-off between the spatial and geometrical information. Given that $k$ is too small (i.e., $k \leq 2$), the spatial information will be largely impaired, which means that the RNN is not able to take full advantage of the sequential data; While if $k$ is too large (i.e., $k \geq 4$), geometrical information becomes vague due to the occurrence of empty slices (with all values equal zero) in the end points of the sequence. For any given 3D shape, voxels centralized in the middle of the volumetric grid, while in the edge region, voxel becomes scattered, which leads to empty integration, as shown in Figure \ref{fig:MDR_n5}. In this case, MDR becomes `scattered', leading to the confusion of the feature generator. Therefore, in our experiment, we choose $k=3$ to make sure that the MDR is both spatial and geometrical informative. As for the sequence organization, we concatenate slices along each axis together to augment the spatial relationship from different views, and an organized sequence with $3k$ elements is formed. 

\textbf{3D-A-Nets Architecture}: 
The proposed network includes two elements: the feature generator $G$, and the classifier $D$. Both elements are trained with Adam optimizer \cite{Kingma2014Adam}, with the batch size 32. Different base learning rates with the same decay rate 0.995 are set for $G$ and $D$. The base learning rate for $G$ is 0.01, while the base learning rate for $D$ is 0.001.
Specifically, $G$ consists of 4 convolution layers, with filters 32, 64, 128, and 256, respectively. Kernel of size $4 \times 4$, with a stride of 2 are used. The four convolution layers are shared for each input segment, and the feature sequence is then fed to a ConvLSTM. Given an input object of size $30 \times 30 \times 30$, $3k$ slices are computed, with the size of $30 \time 30 \times 1$. As a result, the latent space of each slice consists of 256 feature maps of size $2 \times 2$. The latent feature sequence with $3k$ elements is then fed to a ConvLSTM for many-to-one feature extraction. The discriminator $D$ consists of three fully-connected layers of size 128, 64 and $N+1$, respectively, where $N$ is the number of classes.

\subsection{Experimental Tests}

\begin{figure*}[h]
\begin{center}
\includegraphics[width=0.8\linewidth]{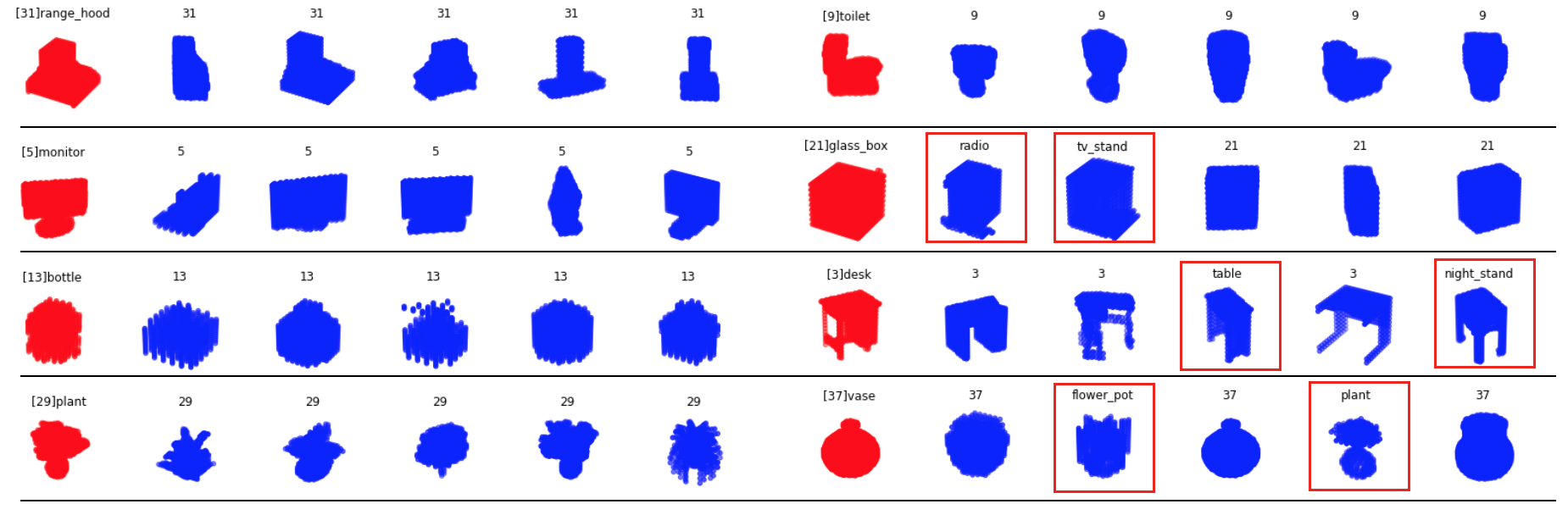}
\end{center}
\caption{Retrieval samples. The left column show query shapes (in red), with identifier and class name, and each query is followed by five shapes with the highest similarity. The wrong retrieval results are marked by red box, with class names labeled.}
\label{fig:retrieval}
\end{figure*}

\textbf{Test 1: Evaluation of MDR for 3D volumetric data}
Multilayer Dense Representation (MDR) is a newly developed hand-crafted representation for 3D data  in this paper. Since 3D-DDSD is learned based upon MDR, in this experiment, we are interested in knowing how the effectiveness of this new representation for 3D volumetric data. To this end, we conduct shape classification on the identical volumetric dataset used by \cite{Wu20153D}, with different method. Differ from the 3D deep belief network employed by \cite{Wu20153D} directly on volumetric data, we first transfer the 3D volumetric representation to the 2D representation (MDR), and then apply 2D convolution networks on MDR. The network consists of 4 Convolution layers, followed by 2 fully-connected layers. The classification accuracy reached 0.856 (as shown in `CNN only' in Table \ref{tab:comparison}), which is better than all other methods with direct application of 3D CNN on 3D volumetric shape. In addition to classification performance, the computation of CNN on 2D dense layer images is much more efficient than that of applying 3D CNN on 3D voxel data directly.

\textbf{Test 2: Evaluation of adversarial learning}
Adversarial learning is an essential component of 3D-A-Nets for the improvement of shape descriptor's quality. We are therefore quite interested in finding out how much performance gained by adversarial learning. To clearly evaluate the effectiveness of the adversarial learning component, we remove the "RNN" from 3D-A-Nets and compare against the methods with CNN-only method. Specifically, we remove the last two fully-connected layers of the original CNN model in 3D shape classification, and take the output of CNN as the latent representation of the segment sequence. The latent representation is then fed to a Discriminator $D$, which aims to recognize all of its inputs as `adversarial'. On the other hand, the feature Generator $G$ fools the $D$ by generating distinguishable features. The classification accuracy reached 0.881 (as shown in `CNN + Adv' in Table \ref{tab:comparison}). We can see from the comparison result that "Adversarial learning" component plays an important role in enhancing the performance of 3D-A-Nets. The significant performance gained with adversarial learning clearly indicates the effectiveness of adversarial learning as an essential component to learn deep shape descriptor in this paper.

\textbf{Test 3: Evaluation of Recurrent Neural Network (RNN)}

RNN component from 3D-A-Nets is fully utilized to exploit the spatiotemporal relationship between the adjacent features generated from MDR. In 3D-A-Nets, ConvLSTM is adopted to better capture spatiotemporal correlations of images than Fully Connected LSTM (FC-LSTM) as claimed in \cite{Shi2015Convolutional}. In this test, we are interested in verifying how much performance that 3D-A-Nets can gain by exploiting spatial relationship among adjacent features. To clearly evaluate the effectiveness of the RNN component, we remove the "adversarial learning" component from 3D-A-Nets and compare against the methods with CNN-only method in 3D shape classification. The classification accuracy reached 0.875 (as shown in `CNN + RNN' in Table \ref{tab:comparison}). We can see from the comparison result that RNN component clearly helps enhance the performance of 3D-A-Nets for representing the 3D shapes. Please note that introducing RNN into the model also leads to an increase in model complexity and computational cost. A carefully chosen number of slices of MDR is of great importance in maintaining a good balance between effectiveness and efficiency. In this paper, we experimentally validate that the selection of 3 slices will maintain a good balance between model complexity and efficiency.

\textbf{Test 4: Evaluation of 3D-A-Nets}

After evaluating on each single component, we integrate all the components together, and verify the overall performance of the proposed 3D-A-Nets model. Compared to single-component evaluation, the proposed 3D-A-Nets achieves higher performance, with classification accuracy 90.5\% and mAP 0.801. Figure \ref{fig:retrieval} illustrates several retrieved examples. The query models are randomly selected from 8 categories, namely, range hood, monitor, bottle, plant, toilet glass, desk and vase, which are are mark with red color and listed at the left most column. The retrieved objects are listed on the right side based on their ranking orders. In most cases, the proposed methods could retrieve correct objects; however, for some particular cases, such as desk and night stand, which are even very difficult for human to distinguish, the proposed methods retrieve wrong objects. All the irrelevant models are marked with red boxes.

\begin{figure}[h]
        \begin{center}
    \includegraphics[width=0.7\linewidth]{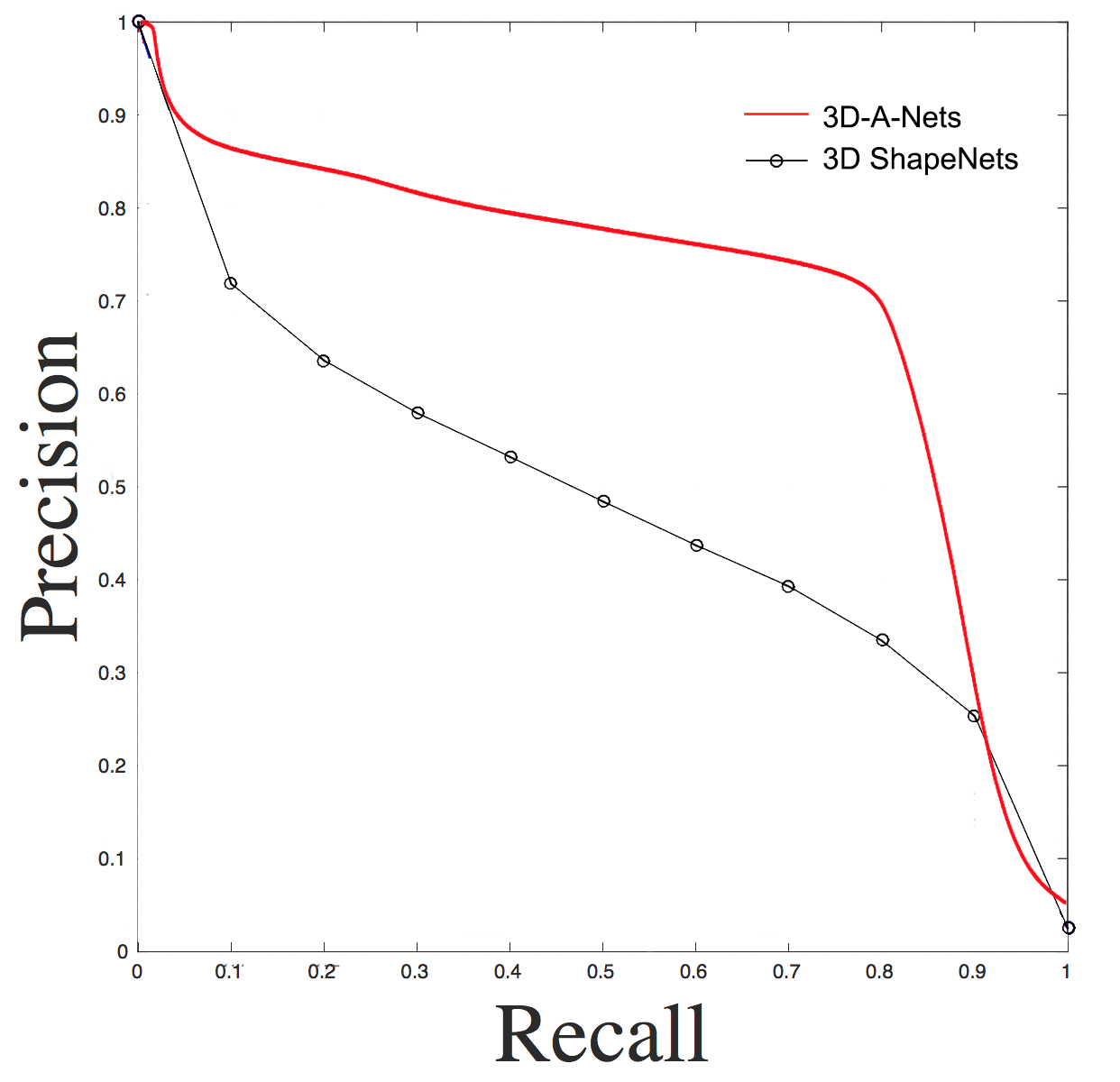}
    \end{center}
        \caption{3D Shape Retrieval on ModelNet40. Precision-recall curves of 3D ShapeNets (black curve) and proposed 3D-A-Nets (red curve) are compared.}
    \label{fig:prc}
\end{figure}

Table \ref{tab:comparison} shows the performance comparison between the proposed methods and the state-of-the-art voxel-based methods, namely 3D ShapeNets \cite{Wu20153D}, 3D-GAN \cite{Wu2016Learning}, and VoxNet \cite{Maturana2015VoxNet}. As shown in the table, the proposed 3D-A-Nets could outperform all the aforementioned methods on Princeton ModelNet40 benchmark with a huge margin. Specifically, the classification accuracy was increased by 7\%, compared to the best state-of-the-arts voxel-based results reported by VoxNet \cite{Maturana2015VoxNet}. Except for quantitative comparisons against state-of-the-art methods, we also draw precision-recall curve to visualize the performance. Figure \ref{fig:prc} shows the precision-recall curve comparison between \cite{Wu20153D} and 3D-A-Nets. The higher curve indicates better performance. As shown in Figure \ref{fig:prc}, the proposed 3D-A-Net is significantly better than 3D ShapeNets.
\begin{table}[h]  
\centering  
\caption{Comparison with the voxel-based methods}
    \begin{tabular}{cccc}
        \hline
        Methods & Classification Acc & mAP \\ [0.5ex] 
        \hline
        3D ShapeNets& 0.77 & 0.492 \\
        3D-GAN & 0.833 & - \\
        VoxNet & 0.83 & - \\
        \textbf{Proposed (CNN only)} & 0.856 & - \\
        \textbf{Proposed (CNN+Adv)} & 0.881 & - \\
        \textbf{Proposed (CNN+RNN)} & 0.875 & - \\
        \textbf{Proposed (3D-A-Nets)} & \textbf{0.905} & \textbf{0.801} \\
        \hline
    \end{tabular}
    \label{tab:comparison}
\end{table}

\section{Conclusion}
% In this work,
% We proposed a unified framework by developing novel MDR shape representation, 
% based on convolutional neural network, recurrent neural network and adversarial nets for learning 3D Deep and Dense shape descriptors with the application in both 3D shape retrieval and classification.
% Our main contributions focus on the generation of the MDR, the utilization of 2D segments on 2D CNN and adversarial learning. The proposed method could aggregate the spatial and sequential relationship of the feature sequence by combining CNN and ConvLSTM. By employing the state-of-the-art techniques from multiple research domains including computational geometry, computer vision and deep learning, the proposed method analyze the complex essence of the 3D model and generate a concise but informative shape descriptor and achieved good classification and retrieval performance.
In this paper, we have developed a novel 3D adversarial network to address the challenging issues of efficient and effective 3D volumetric data processing.
Particularly, in this research, we developed new definition of 2D multilayer dense representation (MDR) to extract concise but geometrically informative shape description and a novel design of adversarial networks that jointly train a set of convolution neural network (CNN), recurrent neural network (RNN) and an adversarial discriminator for the robust 3D-DDSD for volumetric shapes.
% Based on prior discussion and the experiment result, there are still potential improvements to 3D-A-Nets for future work. In experiment settings, the slice number of MDR is determined by trial and error. We tested it from 1 to 5 and finally 3 is chosen.
% However, we have not found the best 
The current parameter of MDR slice number is set up based on experimental experience, and we will develop methods of automatically determining the slice numbers of MDR dynamically based on slice similarity in our future work.
% However, we have found that for certain circumstances, it is possible to gain better performance on more complex 3D shapes by increasing the MDR slice number. If MDR algorithm could adjust the number of slices dynamically according to the variance of all slices or similarities between slices in voxel volume, it may improve the performance for complex 3D shapes and save computation time for less complex 3D shapes. For example, two slices MDR for a coffee table may be good enough but a helicopter may need at least 4 or 5 slices MDR. Smarter MDR generation algorithm may improve the model.

%\pagebreak
%\clearpage
{\small
\bibliographystyle{ieee}
\bibliography{3D_DDSD}
}

\end{document}